\documentclass{article}

\usepackage{PRIMEarxiv}

\usepackage[utf8]{inputenc} 
\usepackage[T1]{fontenc}    
\usepackage{hyperref}       
\usepackage{url}            
\usepackage{booktabs}       
\usepackage{amsfonts}       
\usepackage{nicefrac}       
\usepackage{microtype}      
\usepackage{fancyhdr}       
\usepackage{graphicx}       
\usepackage{xcolor}
\usepackage{mathtools}
\usepackage[square,numbers]{natbib}
\graphicspath{{media/}}     

\newcommand{\projectSource}{\url{https://github.com/radi-cho/GatedTabTransformer}}

\title{The GatedTabTransformer. An enhanced\\
deep learning architecture for tabular modeling.}

\author{
  Radostin Cholakov \\
  High School of Mathematics\\
  Plovdiv, Bulgaria\\
  \texttt{r.cholakov@obecto.com} \\
   \And
  Todor Kolev \\
  Obecto \\
  Sofia, Bulgaria \\
  \texttt{tkolev@obecto.com} \\
}

\begin{document}
\maketitle

\begin{abstract}
    There is an increasing interest in the application of deep learning architectures to tabular data. One of the state-of-the-art solutions is TabTransformer which incorporates an attention mechanism to better track relationships between categorical features and then makes use of a standard MLP to output its final logits. In this paper we propose multiple modifications to the original TabTransformer performing better on binary classification tasks for three separate datasets with more than 1\% AUROC gains. Inspired by gated MLP, linear projections are implemented in the MLP block and multiple activation functions are tested. We also evaluate the importance of specific hyper parameters during training.
\end{abstract}


\section{Introduction}

Some of the most common machine learning pipelines with real-world applications involve manipulation of tabular data. The current state-of-the-art approaches for tabular modeling are treebased ensemble methods such as the gradient boosted decision trees (GBDTs) \cite{chen2016xgboost}. However, there is also an increasing interest in the application of deep learning techniques in the field due to the possibility for bypassing manual embedding creation and feature engineering. \cite{fiedler2021simple}. Multiple neural network solutions such as DNF-Net \cite{abutbul2020dnf}, TabNet \cite{arik1908tabnet} or MLP+ \cite{fiedler2021simple} have been introduced, all of which demonstrate performance comparable to GBDTs.

On the other hand, as we've described in previous studies \cite{cholakov2021transformers}, attention-based architectures, originally introduced to tackle NLP tasks, such as the Transformer \cite{vaswani2017attention} are constantly being adapted to solve a wider range of problems. One proposal is the TabTransformer \cite{Huang2020TabTransformerTD} which focuses on using \textit{Multi-Head Self Attention} blocks to model relationships between the categorical features in tabular data, transforming them into robust contextual embeddings. The transformed categorical features are concatenated with continuous values and then fed through a standard multilayer perceptron \cite{haykin1994neural} (section \ref{sec:tabtransformer}). This way the TabTransformer significantly outperforms pure MLPs and recent deep networks (e.g. TabNet \cite{arik1908tabnet}) for tabular data. We believe that it is possible to further enhance its architecture by replacing the final MLP block with a gated multi-layer perceptron (gMLP) \cite{Liu2021PayAT} - a simple MLP-based network with spatial gating projections, which aims to be on par with Transformers in terms of performance on sequential data (section \ref{sec:gmlp}).

In this paper we will present an enhanced version of the TabTransformer with incorporated gMLP block and the intuition behind it. Also multiple other architecture design decisions based on hyper parameter optimization experiments will be described.

\subsection{The TabTransformer}
\label{sec:tabtransformer}

\begin{figure}[hpt]
  \begin{center}
      \includegraphics[width=10cm]{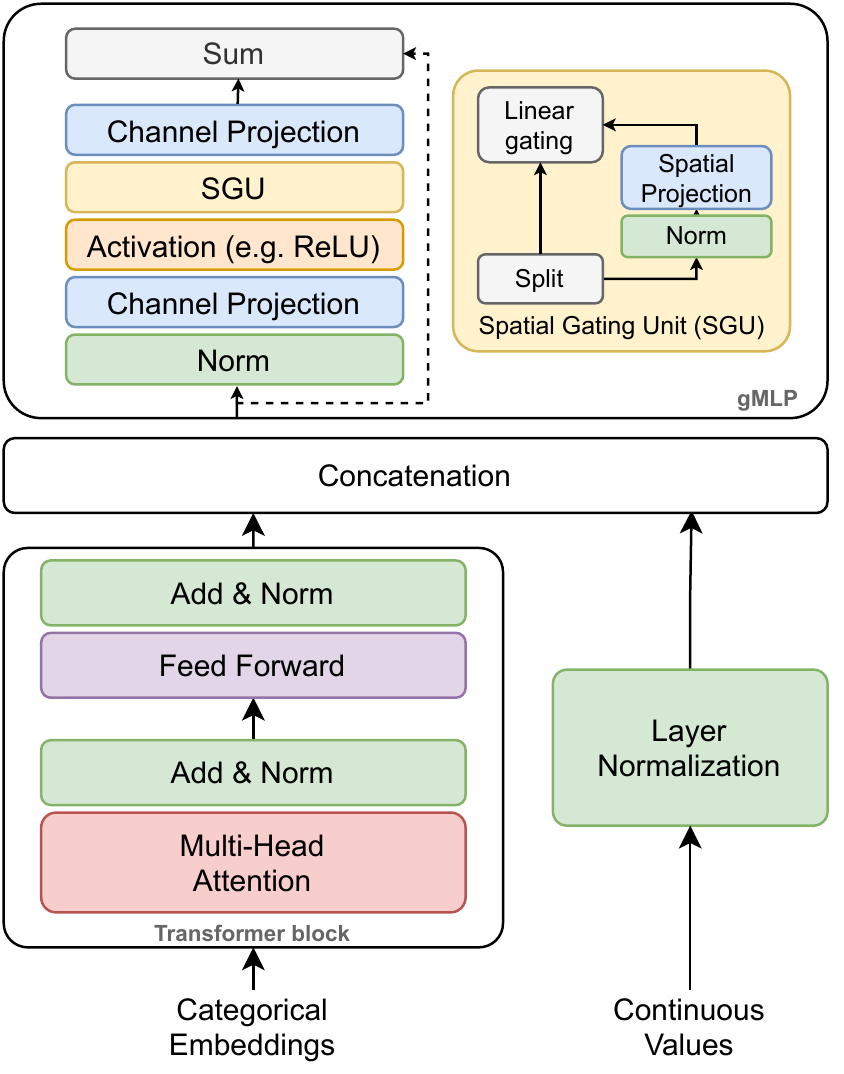}
  \end{center}
  \caption{Architecture of the proposed GatedTabTransformer. $N$ Transformer blocks can be stacked on one another. The same is true for $L$ gMLP layers.}
  \label{fig:model}
\end{figure}

The TabTransformer model, introduced in December 2020 by researchers at Amazon manages to outperform the other state-of-the-art deep learning methods for tabular data by at least 1.0\% on mean AUROC. It consists of a column embedding layer, a stack of $N$ Transformer layers, and a multilayer perceptron (figure \ref{fig:model}). The inputted tabular features are split in two parts for the categorical and continuous values. For each categorical feature the so called \textit{column embedding} is performed (see \cite{Huang2020TabTransformerTD}). It generates parametric embeddings which are inputted to a stack of Transformer layers. Each Transformer layer \cite{vaswani2017attention} consists of a multi-head self-attention layer followed by a position-wise feed-forward layer.

\begin{equation}
    Attention(Q, K, V) = softmax(\frac{QK^T}{\sqrt{d_k}})V
\end{equation}

After the processing of categorical values $x_{cat}$, they are concatenated along with the continuous values $x_{cont}$ to form a final feature vector $x$ which is inputted to a standard  multilayer perceptron.

\subsection{The gMLP}
\label{sec:gmlp}

\begin{figure}[hpt]
    \begin{center}
      \includegraphics[width=9cm]{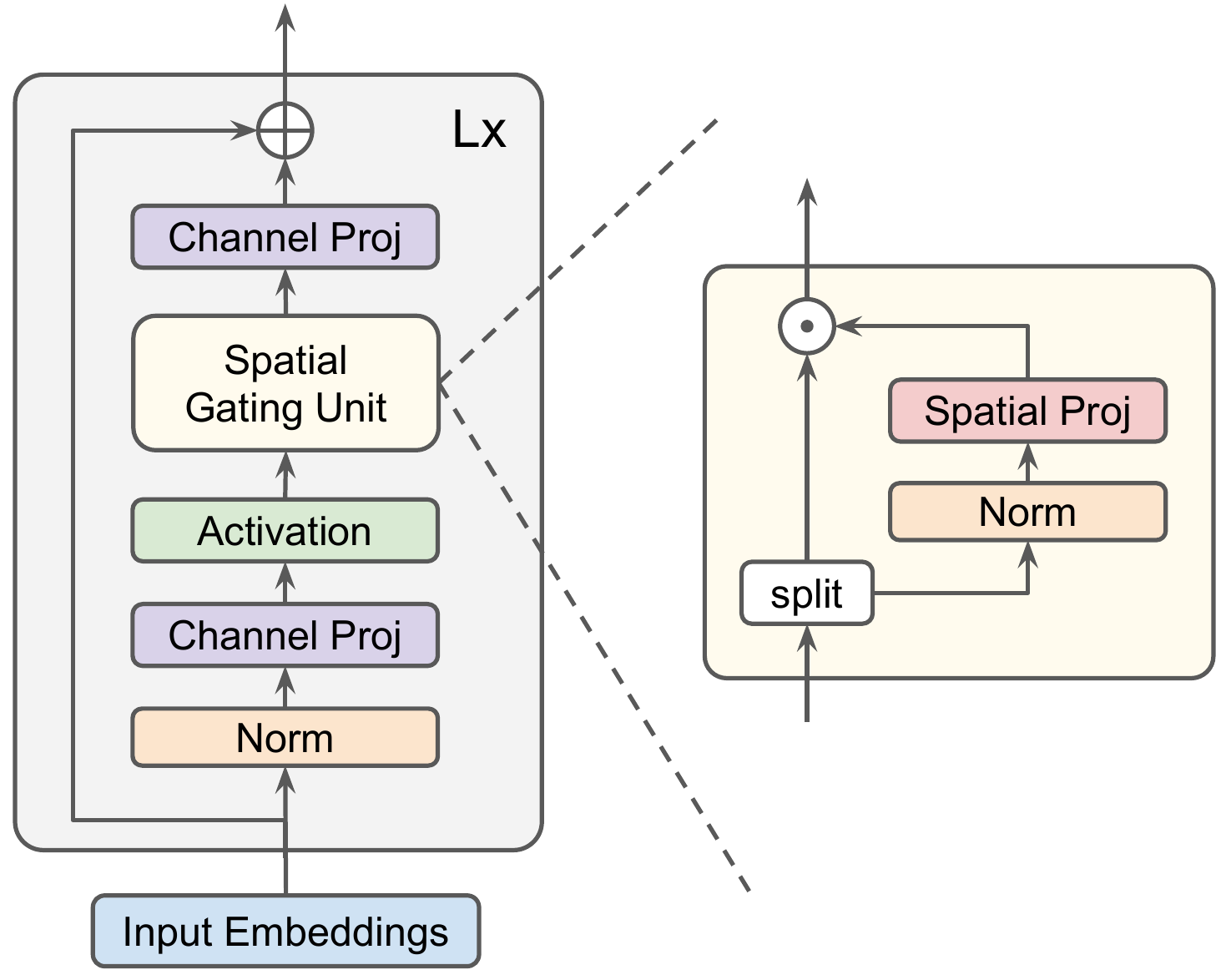}
      \caption{Overview of the gMLP architecture. $Z$ to be split into two independent parts during spatial gating - one for multiplicative bypass and one for gating is also proposed for further optimization.}
    \end{center}

    \label{fig:gmlp-overview}
\end{figure}

The gMLP model \cite{Liu2021PayAT} introduces some simple yet really effective modifications to the standard multilayer perceptron. It consists of a stack of multiple identically structured blocks (figure \ref{fig:gmlp-overview}) defined as follows:

\begin{equation}
    Z = \sigma(XU), \; Z' = s(Z), \; Y = Z'V
\end{equation}

\begin{equation}
    s(Z) = Z * f_{W,b}(Z)
\end{equation}

\begin{equation}
    f_{W,b}(Z) = WZ + b
\end{equation}

where $\sigma$ is an activation function (e.g. ReLU), $U$ and $V$ - linear projections along the channel dimension and $s(.)$ is the so called \textit{spatial gating unit} which captures spatial cross-token interactions. $f_{W,b}$ is a simplistic linear projection and $*$ represents element-wise multiplication. Usually the weights $W$ are initialized as near-zero values and the biases $b$ as ones at the beginning of training.

This structure does not require positional embeddings because relevant information will be captured in the gating units. From the original gMLP paper we can denote that the presented block layout is inspired by inverted bottlenecks which define $s(.)$ as a spatial convolution \cite{sandler2018mobilenetv2}.

gMLP has been proposed by Google in May 2021 as an alternative to Transformers for NLP and vision tasks having up to 66\% less trainable parameters. In our study we will replace the pure MLP block in the TabTransformer with gMLP and test how well it can model tabular data and whether it is able to extract any additional information.

\section{Related work}

In recent years a lot of experiments have been conducted to test the applicability of standard MLPs for tabular modeling \cite{de2015deep}. Multiple Transformer-based methods have been also used to fit tabular features  \cite{li2020interpretable}, \cite{sun2019deepenfm}. For example AutoInt \cite{li2020interpretable} proposes multi-head self-attentive neural network to explicitly model the feature interactions in a low-dimensional space.

An extensive performance comparison of the most popular tabular models along with their advantages and disadvantages can be found in a paper from August 2021 \cite{fiedler2021simple}. It also features enhancements to AutoInt and MLPs such as the use of element-wise linear transformations (gating) followed by LeakyReLU activation. The intuition behind the gates implemented there is similar to the gMLP block in our architecture.

Another recently published paper \cite{kadra2021well} from June 2021, initially named "Regularization is all you Need", suggests that implementing various MLP regularization strategies in combination with some of the already described techniques has a potential for further performance boosts in deep networks for tabular problems.

\section{Model}
\label{sec:model}

As described in section \ref{sec:tabtransformer}, column embeddings are generated from categorical data features. If continuous values are present in the dataset they are passed through a normalization layer.

The categorical embeddings are then processed by a \textit{Transformer block}. In our case a \textit{Transformer block} represents the encoder part of a typical Transformer \cite{vaswani2017attention}. It has two sub-layers -  a multi-head self-attention mechanism, and a simple, positionwise fully connected feed-forward network.

As a final layer we have placed a slight modification of the original gMLP, called \textit{gMLP\_Classification} in our code. It is adapted to output classification logits and works best for optimization of cross entropy or binary cross entropy loss\footnote{https://pytorch.org/docs/stable/generated/torch.nn.BCELoss.html}.

By stacking multiple Transformer blocks and gMLP structures we were able to outperform some state-of-the-art deep models. More details on how well this solution performs can be found in section \ref{sec:results}.

Model implementation\footnote{Model implementation, experiments setup and dataset characteristics are publicly available on \projectSource.} is done with the PyTorch \cite{NEURIPS2019_9015} machine learning framework in a CUDA environment \cite{cuda}.

\section{Experiments}

To test the proposed architecture multiple experiments were conducted. Their main objective was to compare the standard TabTransformer's performance to the proposed model.

\subsection{Data}

We made use of three main datasets for experimentation - namely \textit{blastchar}, \textit{1995\_income} from Kaggle\footnote{\url{https://kaggle.com/}}, and \textit{bank\_marketing} from UCI repository\footnote{\url{http://archive.ics.uci.edu/ml/index.php}}. Their size varies between 7K and 45K samples with 14 to 19 features. In all datasets the label values are binary, thus binary classification should be performed. These three sets are all also mentioned in the original TabTransformer paper \cite{Huang2020TabTransformerTD}. Detailed data characteristics can be found in Appendix \ref{appendix:data}.

Dataset splitting follows the 65/15/20\% pattern for train/validation/test. The validation split is used to pick the best performing model state during training and the test split is only used to determine the final scores.

During our research the \textit{pandas} \cite{mckinney-proc-scipy-2010} and \textit{seaborn} \cite{Waskom2021} packages were used for data analysis, manipulation and visualisation.

\subsection{Hyper parameter optimization}

To generate our results (section \ref{sec:results}) we set up a simplified pipeline without the optional embedding pre-trainning or self-supervised learning steps described in \cite{Huang2020TabTransformerTD}. As a first step we recreated the implementation of the original TabTransformer and fitted it to the mentioned datasets with manually selected hyper parameters. Secondly, tuning environment was created to find the best performing set of hyper parameters (initial learning rate, hidden layers count, hidden dimensions, dropouts, etc.) for each dataset.

Then we upgraded the TabTransformer as described in section \ref{sec:model} and repeated the HPO process again, finding the best set of parameters for the new model. More details on what parameter values were tested can be found in Appendix \ref{appendix:hpo}. To avoid overfitting and to speed up training early stopping was implemented.

To train and tune our models efficiently we used the Ray Tune \cite{liaw2018tune} Python library.

\subsubsection{Learning rate and optimization objectives}

A key parameter when training a ML model is learning rate\footnote{\url{https://en.wikipedia.org/wiki/Learning_rate}}. In our experiments a learning rate scheduler\footnote{\url{https://bit.ly/3lt15km}} was implemented to decay the initial rate $\alpha$ by $\gamma$ every step size $n$ and help for quicker convergence. Values for $\alpha$, $\gamma$ and $n$ together with epoch count patience $p$ for early stopping were tuned with a grid search strategy. The process was executed separately for the gated and baseline TabTransformers.

\subsubsection{Hidden layers and dimensions}

Other important hyper parameters are count and dimensions of the hidden layers in either the MLP or gMLP and number of heads for the \textit{Multi-Head Attention}. All of these were also tuned with Ray. Our findings suggest that by increasing the baseline model's number of MLP neurons its performance peaks at a certain value and slowly decreases from that point onward, whereas the gMLP continues to increase its performance at small steps for much longer. For example if the two models perform equivalently well with hidden dimension of 32, increasing it to 128 is more likely to higher the performance of the gMLP TabTransformer compared to baseline.

\subsubsection{Neuron Activation}

Yet another aspect of optimization is to choose an activation function for the multilayer perceptorn neurons \cite{xu2015empirical}. During tuning we tested ReLU, GELU, SELU and LeakyReLU\footnote{\url{https://paperswithcode.com/method/leaky-relu}} with multiple options for its negative slope (Appendix \ref{appendix:hpo}).

\begin{equation}
    LeakyReLU(x)= 
    \begin{dcases}
        x,& \text{if } x\geq 0\\
        negative\_slope * x,              & \text{otherwise}
    \end{dcases}
\end{equation}

For both models LeakyReLU with slopes from 0.01 up to 0.05 and the standard ReLU performed with highest results. For simplicity and consistency we proceeded our work with standard ReLU activation. More about the effect of LeakyReLU can be found in the following study - \cite{fiedler2021simple}.

\section{Results}
\label{sec:results}

\subsection{Performance evaluation metrics}

For consistency with previously conducted studies we utilized area under receiver operating characteristic curve (AUROC)\footnote{\url{https://en.wikipedia.org/wiki/Receiver_operating_characteristic}} as a performance evaluation technique \cite{bradley1997use}. The reported results are generated by training and evaluating a model configuration with the same parameters but with different seeds or randomized order of data samples multiple times (usually 5, 25 or 50), computing the mean AUROC and then comparing it to the mean scores of the other model configurations. For the TabNet model comparison expected gains were manually calculated based on the results reported in \cite{Huang2020TabTransformerTD} and \cite{fiedler2021simple}.

To estimate and visualise the performance of our models we used the \textit{scikit-learn} \cite{scikit-learn} and \textit{matplotlib} \cite{Hunter:2007} packages for Python.

\subsection{Performance comparisons}

\begin{figure}[hpt]
  \begin{center}
      \includegraphics[width=12.5cm]{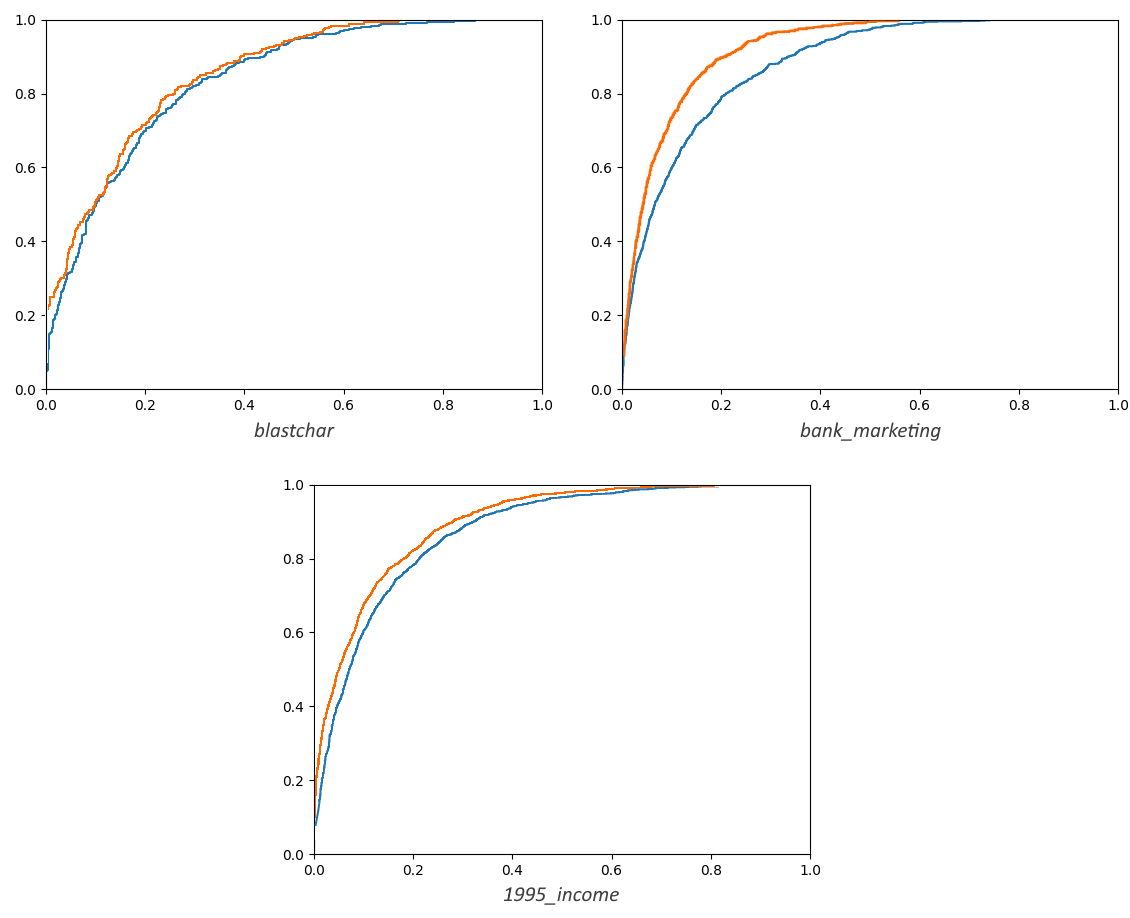}
  \end{center}
  \caption{AUROC gain charts for the 3 datasets - comparison between baseline TabTransformer (\textcolor[rgb]{0.12,0.53,0.7}{blue}) and the proposed GatedTabTransformer (\textcolor[rgb]{1,0.4,0}{orange}).}
  \label{fig:auc-results}
\end{figure}

As a result of our proposals the GatedTabTransformer shows between 0.5\% and 1.1\% performance increase in terms of mean AUROC (figure \ref{fig:auc-results}) compared to the baseline TabTransformer and 1\% to 2\% increase compared to MLPs (table \ref{tab:results-gain}).

\begin{table}[htp]
\caption{
	Performance gain in mean percent AUROC compared to baseline models.
}
\label{tab:results-gain}
\centering
	\begin{tabular}{lrrr}
		\toprule
		        & Gain over & Gain over         & Gain over \\
		Dataset & MLP       & TabTransformer    & TabNet \\
		\midrule
		bank\_marketing  & 1.3 & 1.0 & 3.1 \\
		1995\_income     & 0.9 & 0.7 & 2.5 \\
		blastchar        & 0.4 & 0.5 & 1.6 \\
		\bottomrule
    \end{tabular}
\end{table}

\section{Future work}

A fruitful direction for further research would be to investigate the effects of combining our current solutions with some of the recent breakthroughs in regularization, feature engineering, etc.

As a long term direction we will be working on a custom AutoML\footnote{\url{https://en.wikipedia.org/wiki/Automated_machine_learning}} pipeline focused on tabular data modeling. It will require the implementation of neural architecture search\footnote{\url{https://en.wikipedia.org/wiki/Neural_architecture_search}} strategies such as reinforcement learning or evolutionary algorithms \cite{elsken2019neural} to test and discover new model designs. As a reference a paper titled "The Evolved Transformer" \cite{so2019evolved} by Google from 2019 should be mentioned. It describes how NAS can be used with a Transformer \cite{vaswani2017attention} as initial seed to discover more sophisticated architectures. Analogically TabNet, TabTransformer and other tabular models could be used as seeds in a potential NAS process. Additional embedding techniques and manipulated representations of the data (e.g. TaBERT \cite{yin2020tabert}, TabFormer \cite{padhi2021tabular}) can be incorporated as pre-processing steps.

\section{Conclusion}

In the presented study we have explored some modifications to the original TabTransformer \cite{Huang2020TabTransformerTD} architecture which impact beneficially binary classification tasks for three separate datasets with more than 1\% area under receiver operating characteristic curve gains. Linear projections inspired by gated multilayer perceptrons \cite{Liu2021PayAT} have been proposed for the TabTransformer's final MLP block where its final logits are generated. We have also conducted multiple hyper parameter optimization iterations during training to test the impact of different activation functions, learning rates, hidden dimensions and layer structures. These findings have significant importance when working with tabular predictions and we've open-sourced our model and applied it in practical use cases to further showcase their importance.

\section{Acknowledgements}

This paper is purposed to be a part of the 22-nd Student Conference of the High School Student Institute of Mathematics and Informatics - BAS.

We would like to thank Maria Vasileva and Petar Iliev for their helpful feedback.

\nocite{*}
\bibliographystyle{unsrtnat}
\bibliography{references}

\pagebreak

\appendix
\small
\lhead{Appendix}

\section{HPO parameters}
\label{appendix:hpo}

\begin{itemize}
    \item Learning rates: 0.05, 0.01, 0.005, 0.001, 0.0005
    \item Step sizes for learning rate scheduler: 5, 10, 15 epochs
    \item Learning rate scheduler slope (gamma): 0.1, 0.2, 0.5
    \item Dropout: 0.0, 0.1, 0.2, 0.5
    \item TabTransformer number of heads: 4, 8, 12, 16
    \item MLP/gMLP number of hidden layers (depth): 2, 4, 6, 8
    \item MLP/gMLP dimensions: 8, 16, 32, 64, 128, 256
\end{itemize}

\section{Data description}
\label{appendix:data}

\begin{table}[htp]
\caption{
	Dataset sizes and other details.
}
\label{tab:data-details}
\centering
	\begin{tabular}{lrrrrrrr}
		\toprule
		        &            & Total    & Categorical & Continous & Positive \\
		Dataset & Datapoints & Features & Features    & Features  & Class \% \\
		\midrule
		bank\_marketing  &   $45,211$ &       16 &    11 &    5 &     11.7 \\
		1995\_income     &   $32,561$ &       14 &     9 &     5 &     24.1 \\
		blastchar        &    $7,043$ &       19 &    17 &     2 &     26.5 \\
		\bottomrule
    \end{tabular}
\end{table}

\begin{table}[htp]
\caption{Dataset sources. From \cite{Huang2020TabTransformerTD}.}
\label{tab:dataset-urls}
\centering
\scalebox{0.95}{
\begin{tabular}{ll}
\toprule
     Dataset Name &  URL \\
\midrule
     1995\_income & \url{https://www.kaggle.com/lodetomasi1995/income-classification} \\
  bank\_marketing & \url{https://archive.ics.uci.edu/ml/datasets/bank+marketing} \\
        blastchar & \url{https://www.kaggle.com/blastchar/telco-customer-churn} \\
\bottomrule
\end{tabular}
}
\end{table}

\begin{figure}[hpt]
  \begin{center}
      \includegraphics[width=10.5cm]{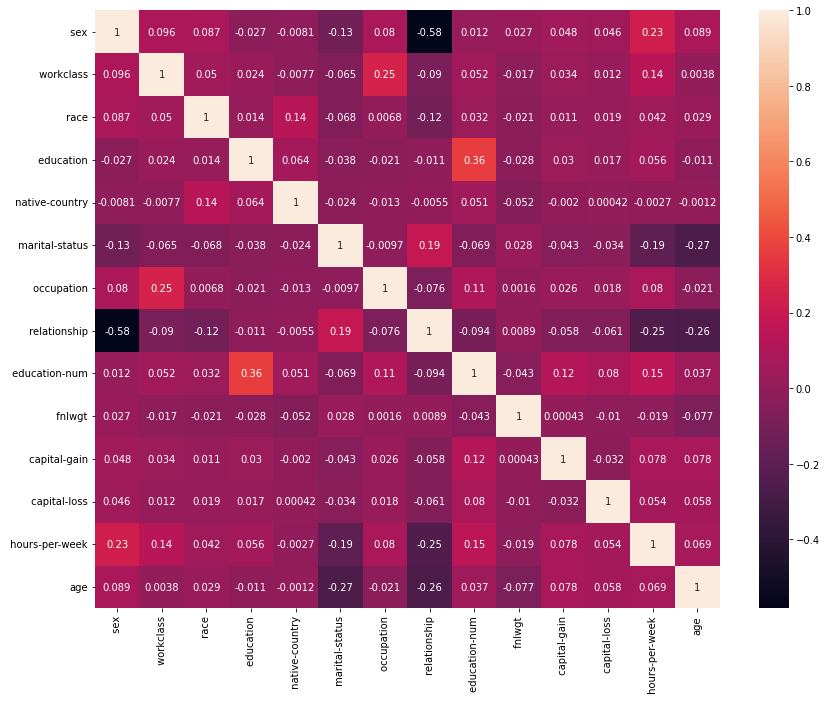}
  \end{center}
  \caption{Correlation matrix for the \textit{1995\_income} dataset.}
  \label{fig:correlation-income}
\end{figure}

\begin{figure}[hpt]
  \begin{center}
      \includegraphics[width=11.5cm]{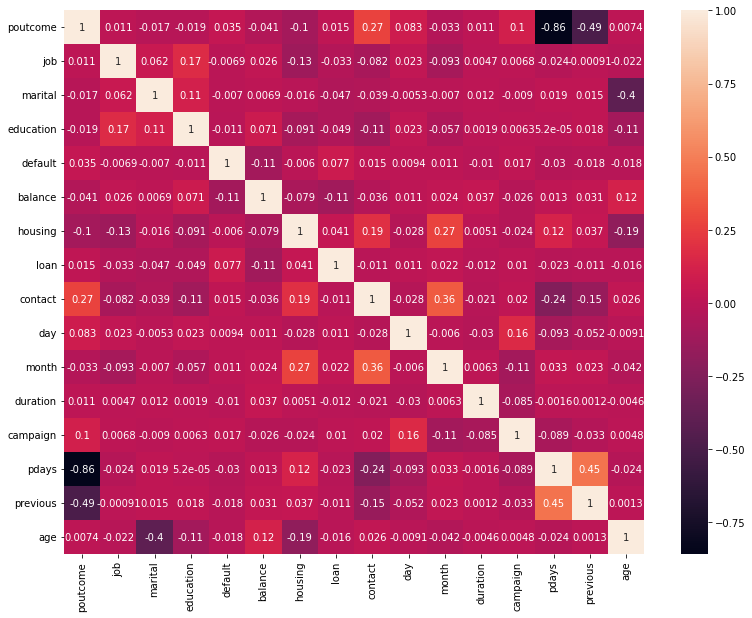}
  \end{center}
  \caption{Correlation matrix for the \textit{bank\_marketing} dataset.}
  \label{fig:correlation-bank}
\end{figure}

\begin{figure}[hpt]
  \begin{center}
      \includegraphics[width=12.5cm]{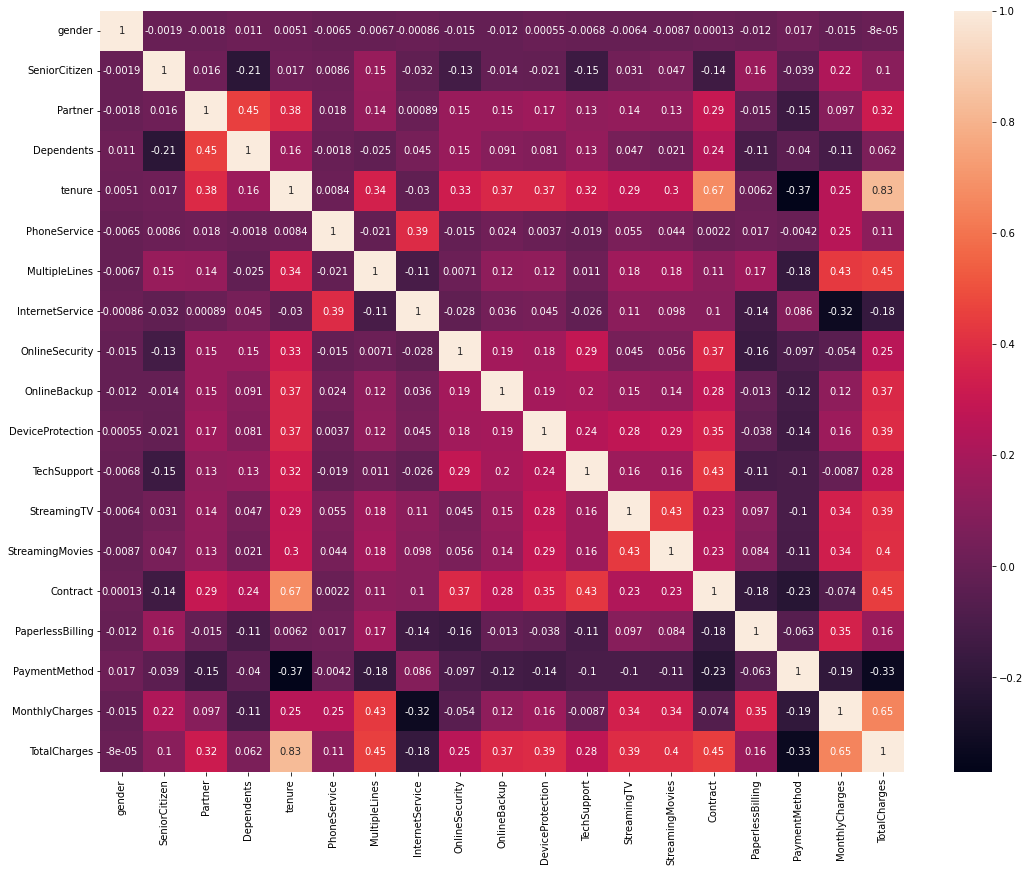}
  \end{center}
  \caption{Correlation matrix for the \textit{blastchar} dataset.}
  \label{fig:correlation-blastchar}
\end{figure}

\end{document}